\documentclass[lettersize,journal]{IEEEtran}
\ifCLASSOPTIONcompsoc
\usepackage[nocompress]{cite}
\else
\usepackage{cite}
\fi
\usepackage{algorithmic}
\usepackage{array, tabularray}
\usepackage{textcomp}
\usepackage{stfloats}
    
\usepackage{xcolor}
\usepackage{tikz}
\usetikzlibrary{external}
\usepackage{pgfplots}
\pgfplotsset{compat=1.18}

\usepackage{verbatim}
\usepackage{algorithm,amsbsy,amsmath,amsfonts,amssymb,epsfig,bbm,mathrsfs,multirow,amsthm,xcolor,cite}
    
\usepackage{epstopdf}
\usepackage[hidelinks]{hyperref}
\usepackage{graphicx,caption,subcaption}
\usepackage{setspace}
\usepackage[labelformat=simple]{subcaption}

\PassOptionsToPackage{hyphens}{url}
\usepackage{url}

\usepackage{hyperref}
\hypersetup{
    colorlinks=true,
    linkcolor=blue,
    filecolor=magenta,      
    urlcolor=cyan,
    citecolor=blue,
    }
    
\begin{document}

\title{Privacy-Preserving Driver Drowsiness Detection with Spatial Self-Attention and Federated Learning}


\author{Tran Viet Khoa, Do Hai Son, Mohammad Abu Alsheikh, Yibeltal F Alem, and Dinh Thai Hoang	
        
\thanks{T.~V.~Khoa, M. Abu Alsheikh and Y. F. Alem are with the University of Canberra, Australia (e-mail: \{khoa.tran, mohammad.abualsheikh, yibe.alem\}@canberra.edu.au).} 

\thanks{D.~H.~Son is with the School of Electrical Engineering, Computing and Mathematical Sciences, Curtin University, Australia and the VNU Information Technology Institute, Hanoi, Vietnam (e-mail: dohaison1998@vnu.edu.vn).}

\thanks{D.~T.~Hoang is with the School of Electrical and Data Engineering, University of Technology Sydney, Australia (e-mail: hoang.dinh@uts.edu.au).}
}
\maketitle

\noindent\textit{\textbf{Note:} This work has been submitted to the IEEE for possible publication. Copyright may be transferred without notice, after which this version may no longer be accessible.} \\

\begin{abstract}
Driver drowsiness is one of the main causes of road accidents and is recognized as a leading contributor to traffic-related fatalities. However, detecting drowsiness accurately remains a challenging task, especially in real-world settings where facial data from different individuals is decentralized and highly diverse. In this paper, we propose a novel framework for drowsiness detection that is designed to work effectively with heterogeneous and decentralized data. Our approach develops a new Spatial Self-Attention (SSA) mechanism integrated with a Long Short-Term Memory (LSTM) network to better extract key facial features and improve detection performance. To support federated learning, we employ a Gradient Similarity Comparison (GSC) that selects the most relevant trained models from different operators before aggregation. This improves the accuracy and robustness of the global model while preserving user privacy. We also develop a customized tool that automatically processes video data by extracting frames, detecting and cropping faces, and applying data augmentation techniques such as rotation, flipping, brightness adjustment, and zooming. Experimental results show that our framework achieves a detection accuracy of 89.9\% in the federated learning settings, outperforming existing methods under various deployment scenarios. The results demonstrate the effectiveness of our approach in handling real-world data variability and highlight its potential for deployment in intelligent transportation systems to enhance road safety through early and reliable drowsiness detection.

\end{abstract}


\section{Introduction}\label{sec:introduction}

Technological advancements in transportation have significantly improved road safety and driving efficiency, particularly in autonomous and semi-autonomous vehicles. Various driver assistance systems, including fatigue detection technologies, have been developed to mitigate risks associated with human error. Despite these improvements, driver fatigue remains a critical factor in road accidents, contributing to 10\%–20\% of serious crashes worldwide~\cite{thomas2021fatigue}. In Australia, fatigue is recognized as one of the ``fatal five'' causes of road accidents, alongside speeding, drug and alcohol impairment, failure to wear seatbelts, and driver distraction~\cite{thomas2021fatigue}.

There are two major approaches to detecting drowsiness to reduce these risks: physiological signals and vision-based methods~\cite{fu2024survey}. Physiological signals, such as Electroencephalogram (EEG)~\cite{yang2019complex, reddy2021eeg} and Electrocardiogram (ECG)~\cite{fujiwara2023driver}, rely on biological signals, including brain waves and heart rate variations, to identify specific biological markers of drowsiness. However, this approach requires complex systems to record an individual’s biosignals, making it suitable for lab environments but challenging to deploy in practical settings~\cite{fu2024survey}. In contrast, vision-based drowsiness detection~\cite{garcia2012vision} uses visual cues, such as head position, eye movements, and mouth activities, to assess driver fatigue. Effective drowsiness detection must identify subtle signs of drowsiness, such as small changes in eye behavior or facial expressions, while adapting to individual differences and varying in-vehicle conditions. This approach is more user-friendly and widely applied in various domains, including in-vehicle driver assistance~\cite{tran2018human} and driver fatigue monitoring~\cite{zhang2022systematic}. Additionally, recent advances in machine learning have enabled the development of robust, real-time fatigue detection systems aimed at enhancing driver safety and preventing accidents~\cite{perkins2022challenges}. However, the accuracy of vision-based methods can be influenced by factors such as background complexity, video quality, and dataset heterogeneity across different individuals~\cite{fu2024survey}. Moreover, since drowsiness-related data is inherently distributed across various locations, federated learning offers a promising solution. It enables accurate detection in a decentralized setting while preserving data privacy and minimizing network overhead by avoiding the transfer of large datasets~\cite{singh2025cognitive}.

There are several challenges in data distribution in vision-based drowsiness detection systems. First, due to the nature of drivers' faces dataset, each individual's face is unique, leading to heterogeneous data~\cite{chellapandi2023federated, singh2025cognitive}. This data can cause deep learning models to fall into local optima, preventing convergence and reducing accuracy~\cite{chen2024lightweight}. This challenge becomes more significant in decentralized environments, where analyzing data requires combining knowledge from multiple deep learning models. 
Second, in federated learning, the overall system accuracy can be compromised if individual clients contribute low-quality models that have learned patterns significantly different from those of other clients~\cite{singh2025cognitive}. Such inconsistencies may lead to disruptive updates during model aggregation at the central server, ultimately degrading the performance of the global model. This challenge is particularly relevant in drowsiness detection, where variations in driver behavior, environment, and recording conditions across clients can result in highly divergent local models~\cite{singh2025cognitive}.
Third, existing datasets are often limited in size and diversity, failing to capture the full range of real-world driving conditions. This lack of representativeness can hinder the generalizability of drowsiness detection models. Moreover, each video frame includes complex background environments, and variations in lighting, particularly between day and night, can significantly degrade detection accuracy~\cite{shaik2023systematic}.


To deal with the first challenge of handling heterogeneous facial data, our proposed framework employs SSA to emphasize the most important features of each extracted face. By applying an SSA mechanism, the model can focus on regions such as the eyes and mouth, which are critical for drowsiness detection. This approach reduces variations between images, ensuring more consistent data. By emphasizing key facial features, SSA improves the model’s ability to adapt to different conditions, resulting in more reliable and accurate drowsiness detection.
To address the second challenge, we implement GSC in the federated learning model to filter out dissimilar learned knowledge from operators. This ensures the quality of local models while maintaining the accuracy and robustness of the aggregated global model. The GSC identifies which updates are most aligned with the global learning objective, ensuring that only meaningful updates contribute to model improvement. By selecting gradients across operators, this approach enhances model convergence and reduces the impact of noisy or biased updates, leading to more stable and reliable learning outcomes.   
Finally, to address the third challenge, we build a customized frame extraction and augmentation tool that automatically extracts frames from videos. It can also perform face detection and extraction to remove the background elements of the images. After that, we deploy augmentation techniques to generate variations of the original images, enhancing the dataset. This process not only increases the diversity of the data but also improves the model's ability to recognize faces under different conditions, ensuring more robust performance in real-world applications. Our contributions can be summarized as~follows:

\begin{itemize}
    \item We propose a novel framework for detecting driver drowsiness that can effectively handle heterogeneous data. This framework is capable of functioning in decentralized environments, ensuring its applicability in real-world scenarios where data is often distributed and privacy concerns are paramount.

    \item We develop a preprocessing tool that extracts frames from raw video streams, performs face detection and cropping, and applies frame augmentation techniques to increase dataset variability and model generalization. 
    
    \item We develop a highly effective federated learning-based model integrating with GSC at the server side to select the appropriate models from operators for aggregation. In addition, our model employs SSA and LSTM at local training operators to improve the accuracy of detection. This approach ensures that the global model remains accurate and robust, even when working with decentralized and diverse data sources, while also preserving data privacy.

    \item We perform extensive simulations in a real-world dataset to evaluate our system. The results show that our model outperforms existing methods in both centralized and federated learning. Our proposed model can achieve up to 89.9\% accuracy with federated learning. Additionally, our experiments demonstrate that the model can easily adapt to new participants without prior training, making it practical for real-world applications. 
\end{itemize}

\section{Related works}\label{sec:related_works}

There are several works trying to deal with detecting drowsiness using computer vision. In~\cite{mittal2021driver}, the authors use various machine learning and deep learning models (i.e., K-Nearest Neighbour, Naïve Bayes, Logistic Regression, Decision Trees, Random Forest, XGBoost, MLP, and CNN) to compare their performance in drowsiness detection in the University of Texas at Arlington Real-Life Drowsiness Dataset (UTA-RLDD). The simulation results show that the Logistic Regression achieves the highest accuracy of up to 75.67\%.
In~\cite{mou2021isotropic}, the authors introduce an isotropic self-supervised
learning with momentum contrast (IsoSSL-MoCo) model to learn the representations of participants' images and exploit the complementarity of multimodal data. They propose a fusion model that is pretrained by the IsoSSL-MoCo to improve the performance of driver drowsiness detection. The simulation results with the NTHU-DDD dataset show that their proposed solution can achieve an accuracy of up to 93.71\%. 
In~\cite{pandey2023dumodds},  the authors propose two models for detecting drowsiness from a dataset. The first model (Model-A) combines YOLOv3~\cite{redmon2018yolov3} and LSTM, while the second model (Model-B) integrates CNN and LSTM. The simulation results show that although Model-A is more complex than Model-B, it achieves a lower accuracy of 86\% compared to Model-B’s 97.5\%. However, Model-A offers advantages in training efficiency over Model-B. In~\cite{usmani2023driver}, the authors propose using Vision Transformers (ViT) for driver drowsiness detection. The simulation results show that their approach achieves a test accuracy of up to 98.10\% and an average prediction time of approximately 17 ms per frame. In~\cite{krishna2022vision}, the authors propose an approach that pretrains the dataset using YOLOv5 to detect and extract participants' faces. After that, Vision Transformers are employed to detect drowsiness. The simulation results show that their approach can achieve an accuracy of up to 95.5\%. 

All the above methods focus on centralized learning where all data is collected into a central server for analysis. However, in practice, due to the nature of decentralisation of car-driving environment, it is difficult to gather all data into a centralized server without the risk of compromising data privacy. In~\cite{zhao2023fedsup}, the authors propose a federated learning approach for detecting fatigue driving behaviors. Their method uses edge servers to manage client data, while federated learning on cloud servers aggregates the learned knowledge from these edge servers. The proposed model, FedSup, enhances model sharing efficiency and reduces communication overhead. The simulation results show that their approach achieves an accuracy of approximately 90\% in detecting fatigue driving~behaviors. 

We observe that many existing methods formally divide data into training and testing datasets. However, one of the biggest limitations in many drowsiness detection studies is that the same participants often appear in both training and testing datasets~\cite{Cui2023}. Specifically, all of the aforementioned methods create training and testing datasets using different features from the same participants.
This thus allows the learning model to learn individual-specific patterns, such as unique facial features or eye movement behaviors. As a result, the reported performance may not reflect how well the model works on completely new users.
Therefore, in this paper, we consider a more practical scenario where a trained model can be used to detect drowsiness for new participants who have not been previously included in the training process. 
Recently, the authors in~\cite{zhang2022privacy} propose to use separating participants in the training and testing datasets. They also employ federated learning for fatigue detection in a decentralized driving environment. However, their primary focus is on evaluating the impact of noise in federated learning to enhance the privacy of drivers' data. Although their approach uses different participants for training and testing,  it achieves limited performance, with a maximum accuracy of approximately 70\% for drowsiness detection.
In this paper, we also consider a decentralized driving environment while evaluating the performance of participant separation in training and testing datasets, but with a focus on improving detection accuracy while maintaining data privacy. To achieve this, our model is designed to easily adapt to new participants for drowsiness detection. This adaptability enhances its practicality for real-world applications, where new users may continuously join the system. Extensive simulation results show that our proposed model achieves an accuracy of 89.9\% on the testing dataset which includes both trained and untrained participants.
\section{Proposed drowsiness detection Framework}\label{sec:drowsiness_detection}
\begin{figure*}[t!]
    \centering
	\includegraphics[width=0.8\linewidth]{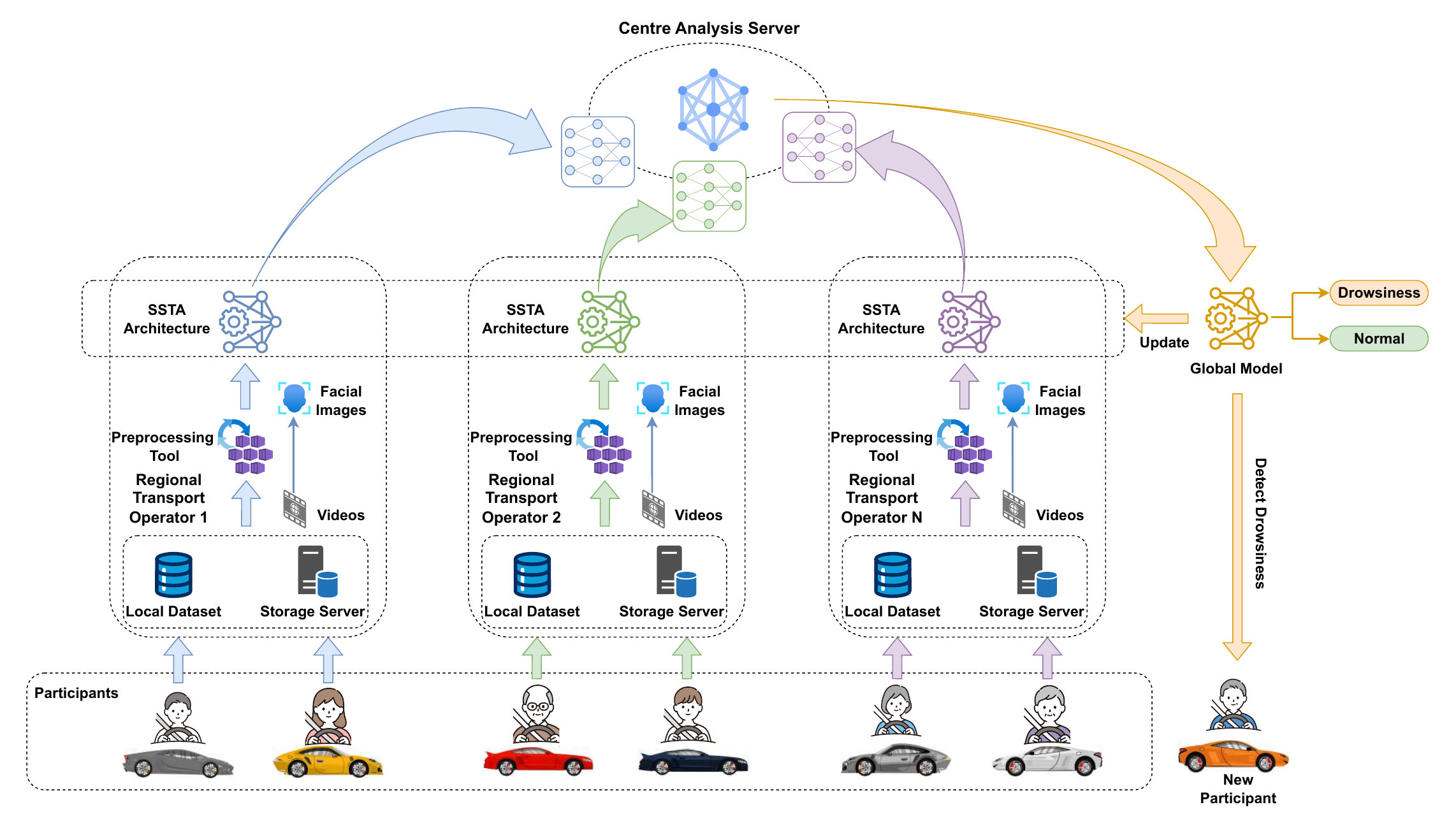}
	\caption{The proposed system model.}
	\label{fig:System}
\end{figure*}

In this paper, we propose a framework for detecting drowsiness in a decentralized car-driving environment. Fig.~\ref{fig:System} describes our proposed system model. As shown in Fig.~\ref{fig:System}, there are $N$ regional transport operators (operators), each collecting data from local participants who may vary in age, gender, and background environment. 

In practice, video recordings of participants will be collected during the experiment and subsequently reviewed by trained operators for labeling. Each operator will carefully analyze the videos to identify and classify the participants' states of alertness, marking specific time segments where signs of drowsiness are observed. In particular, transport operator experts can review video recordings of drivers and tag time segments as drowsy or alert based on clear observable cues. For example, they can mark a period as drowsy whenever the driver shows obvious fatigue indicators, e.g., prolonged eye closures (extended blinks beyond the normal ~0.1 -0.4 second range)~\cite{jimaging9050091}, frequent yawning, or episodes of head nodding where the driver's head briefly droops~\cite{yang2022fatigueview}. To keep the labeling consistent and objective, the operators can follow standardized guidelines, e.g., defining any blink longer than a certain threshold (e.g., 0.4 seconds) as a drowsiness event~\cite{jimaging9050091},  so that each video is judged by measurable behaviors rather than personal guesswork. This labeling process is designed to be feasible in real operational settings, i.e., it relies only on regular camera footage and human observation, without any specialized medical instruments or clinical tests.


Since facial videos are sensitive data, they are not shared across operators or networks. Furthermore, because each operator only has access to a limited amount of participant data, it becomes challenging to train accurate models locally. To address these issues, our framework enables collaborative model training across the operators while preserving user privacy. This allows operators to improve the performance of drowsiness detection models without sharing raw data.
Each operator has a storage server for managing its local dataset, ensuring that sensitive data remains private and is not transmitted across the network. Within each operator, to prepare this data for training and detection, a preprocessing tool is used to convert the recorded videos into sequences of facial images. This includes steps such as face detection, face extraction, and data augmentation, which are explained in detail in Section~\ref{sec:preprocessing}.  After preprocessing, a detection module based on SSA and LSTM is used to identify signs of drowsiness in facial image sequences. More details about this processing architecture can be found in Section~\ref{sec:SSTA_processing}.


Due to the limited training data available in each operator, it is essential for operators to exchange learned knowledge to enhance detection accuracy. To address this, each operator sends its trained model to a central analysis server (e.g., the National Transport Authority). The central analysis server uses the trained models from operators to compute gradients, selects the most valuable information through gradient selection, and aggregates it into a new global model.
This global model is then used to update the deep learning models within each regional transport operator, enhancing their ability to detect drowsiness. More importantly, the updated model can also be used by new participants to detect drowsiness while driving, improving the general adaptability and effectiveness of the system. This decentralized learning strategy is also used in real-world, large-scale systems. For instance, Tesla's Autopilot system adopts a fleet learning approach, where each vehicle processes driving data locally and sends only summarized model updates to a central server. This enables the global model to improve continuously while ensuring that raw video and sensor data remain on the vehicle~\cite{tesla_fleet_learning, Tesla_Autopilot}.

Overall, in this paper, we propose a novel framework that can learn from different groups of people's videos to detect drowsiness with high accuracy while preserving privacy. Our proposed framework includes three main processes as follows: the Preprocessing Process, the SSA and Temporal Aggregation network (SSTA) architecture, and the Federated Learning Drowsiness Detection.


\subsection{Preprocessing Process}\label{sec:preprocessing}
\begin{figure*}[!h]
    \begin{center}
    \includegraphics[width=0.8\linewidth]{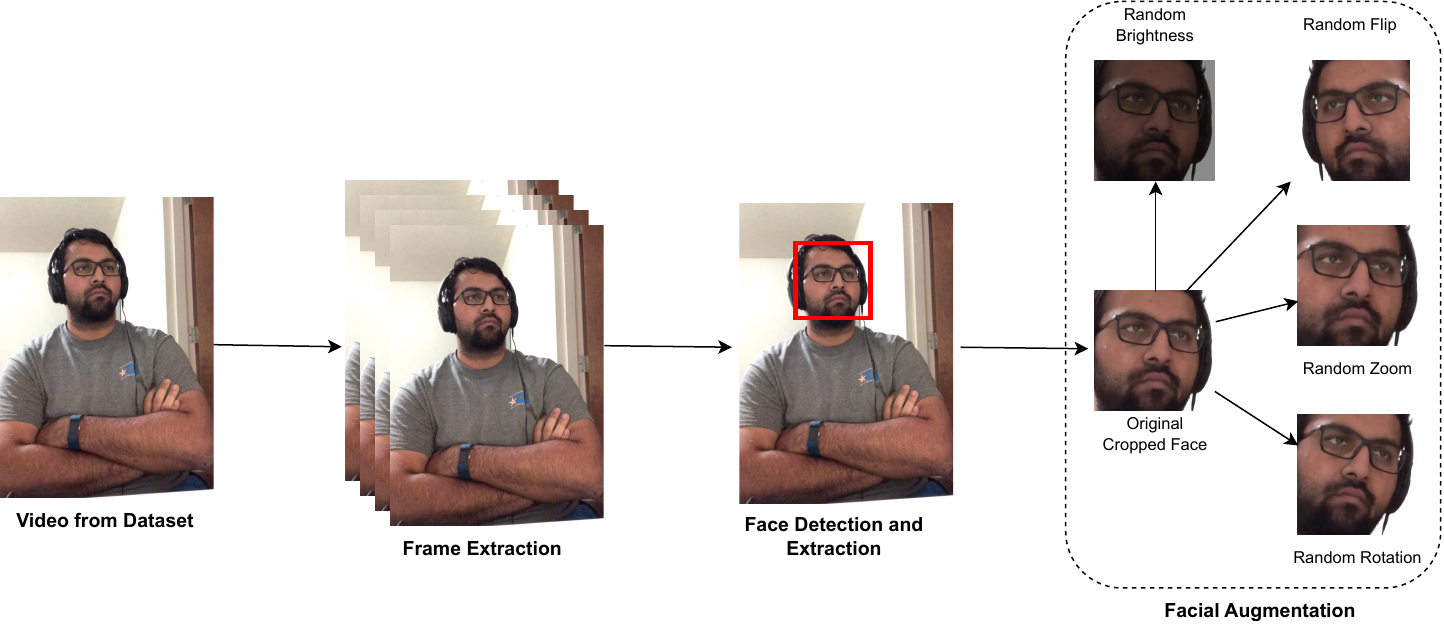}
	\caption{Overview of the face recognition tool, which processes video through three stages: frame extraction, face detection (to isolate user faces and remove backgrounds), and facial augmentation to enhance dataset diversity.}
    \label{fig:preprocessing}
    \end{center}
\end{figure*}

We develop a tool that preprocesses video data by extracting multiple frames from videos, detecting and extracting faces, and augmenting frames. The processes of this tool are described in Fig.~\ref{fig:preprocessing}. In the first step, the tool first extracts the video into multiple frames within a predefined time window and then performs the next processes as follows.

\subsubsection{Face Detection and Extraction}
First, we detect and extract the face from a frame to support the processing model in the next step, which focuses on detecting user drowsiness. To do that, we integrate the face recognition framework~\cite{face-recognition} into our tool. This framework uses the Histogram of Oriented Gradients (HoG)~\cite{dalal2005histograms} to capture the gradient structure of a frame for object detection, e.g., faces. HoG operates by computing gradient orientations and magnitudes over a frame. Denoting a frame as $F(x,y)$, the gradients in the horizontal direction $G_x(x,y)$ and vertical direction $G_y(x,y)$ can be calculated as follows~\cite{gonzalez2002digital}:
\begin{equation}
\begin{aligned}
\label{eqn1}
G_x(x, y) = \frac{\partial F(x,y)}{\partial x}, \quad G_y(x, y) = \frac{\partial F(x,y)}{\partial y},
\end{aligned}
\end{equation}
where $x$ and $y$ are pixel positions of a frame. The gradient magnitude $M(x, y)$ and orientation $\epsilon(x, y)$ are then calculated as follows~\cite{gonzalez2002digital}:
\begin{equation}
\begin{aligned}
\label{eqn2}
M(x, y) &= \sqrt{G_x(x, y)^2 + G_y(x, y)^2}, \\
\epsilon(x, y) &= \tan^{-1} \left( \frac{G_y(x, y)}{G_x(x, y)} \right).
\end{aligned}
\end{equation}

The frame is then divided into small spatial regions called cells, where the gradient orientations are quantized into bins, and a histogram of these orientations is built. The bin $E_j$ (the value of the $j$-th orientation range in a histogram) is updated as follows~\cite{dalal2005histograms}:
\begin{equation}
\begin{aligned}
\label{eqn3}
E_j= \sum_{(x,y) \in \text{cell}}M(x, y) \,\psi(\epsilon(x, y),j),
\end{aligned}
\end{equation}
where $\psi(\epsilon,j)$ is a weighting function that distributes the gradient magnitude into adjacent bins based on linear interpolation. Each cell thus produces a histogram $\mathbf{E}^{\text{(cell)}} = [E_1, E_2, \dots, E_c]$, where $c$ is the number of orientation bins per cell histogram. The histograms from cells are grouped into spatial regions called blocks. Suppose each block contains $a$ cells. Then, for block $b$, the concatenated histogram vector is constructed by stacking the cell histograms~\cite{dalal2005histograms}:
\begin{equation}
\begin{aligned}
\label{eqnV}
\mathbf{V}^{(b)} = [E_1^{(b)}, E_2^{(b)}, \dots, E_{ac}^{(b)}] \in \mathbb{R}^{ac},
\end{aligned}
\end{equation}
where \( E_k^{(b)} \) represents the $k$-th bin from the collection of cell histograms within block $b$. To improve robustness against illumination changes, block normalization is applied using the L2-norm~\cite{dalal2005histograms}:
\begin{equation}
\begin{aligned}
\label{eqn4}
\mathbf{V'}^{(b)} = \frac{\mathbf{V}^{(b)}}{\sqrt{\|\mathbf{V}^{(b)}\|^2 + \eta^2}},
\end{aligned}
\end{equation}
where $\eta$ is a small constant to avoid division by zero. Finally, the normalized vectors from all blocks are concatenated to form the global HoG feature descriptor~\cite{dalal2005histograms}:
\begin{equation}
\begin{aligned}
\label{eqn5}
\mathbf{F'} = [\mathbf{V'}^{(1)}, \mathbf{V'}^{(2)}, \dots, \mathbf{V'}^{(L)}] \in \mathbb{R}^{Lac},
\end{aligned}
\end{equation}
where $L$ is the total number of blocks in the image. Finally, a Support Vector Machine (SVM) classifier is used to distinguish between face and non-face regions based on the extracted HoG features~\cite{dalal2005histograms}. Based on the HoG feature descriptor $\mathbf{F'}$ and the SVM classifier, the facial region is located and extracted from the original input frame. Let us denote the original image frame as $F_{\text{img}}$. The cropped face image used for further processing is then defined as:
\begin{equation}
\begin{aligned}
\label{eqn_face_crop}
I' = \text{Crop}(F_{\text{img}}),
\end{aligned}
\end{equation}
where $\text{Crop}(\cdot)$ is a function that extracts the detected face region from the input frame $F_{\text{img}}$. This image $I'$ serves as the input to the facial augmentation process described in the next subsection. Fig.~\ref{fig:preprocessing} presents an illustration of the face detection implemented in this tool.

\subsubsection{Facial Augmentation}

In vision-based object detection, increasing the amount of training data is crucial for improving model accuracy. Facial augmentation is a widely used technique that artificially expands the dataset by applying a series of transformations to the original facial images. Such transformations include flipping, rotation, scaling, cropping, and adjustments to brightness, contrast, or saturation.

Formally, let $I' \in \mathbb{R}^{H \times W \times C}$ denote an input image, where $H$, $W$, and $C$ represent the height, width, and number of color channels, respectively. With $Z$ representing the total number of different augmentation transformations applied to the original image $I'$, the facial augmentation applies a set of transformation functions $\mathcal{T} = \{T_0, T_1, T_2, \dots, T_Z\}$ to generate augmented images:
\begin{equation}
\begin{aligned}
\label{eqn_F1}
I_z = T_z(I'), \quad z = 0, 1, 2, \dots, Z,
\end{aligned}
\end{equation}
where each $T_z$ represents a specific augmentation operation (e.g., rotation or flipping), and $z=0$ means no transformation applied to the original facial image. The augmented dataset is thus composed of the original facial image along with its transformed variants:
\begin{equation}
\begin{aligned}
\label{eqn_F2}
\mathcal{D}_{\text{aug}} = \{I_0, I_1, I_2, \dots, I_z\}.
\end{aligned}
\end{equation}

By creating diverse variations of the original data, the facial augmentation helps the model generalize better across different scenarios, thereby enhancing robustness and accuracy during training. Fig.~\ref{fig:preprocessing} illustrates the augmentation process applied to a cropped face image.


\subsection {Proposed SSTA Architecture}\label{sec:SSTA_processing}
\begin{figure*}[t!]
	\includegraphics[width=\linewidth]{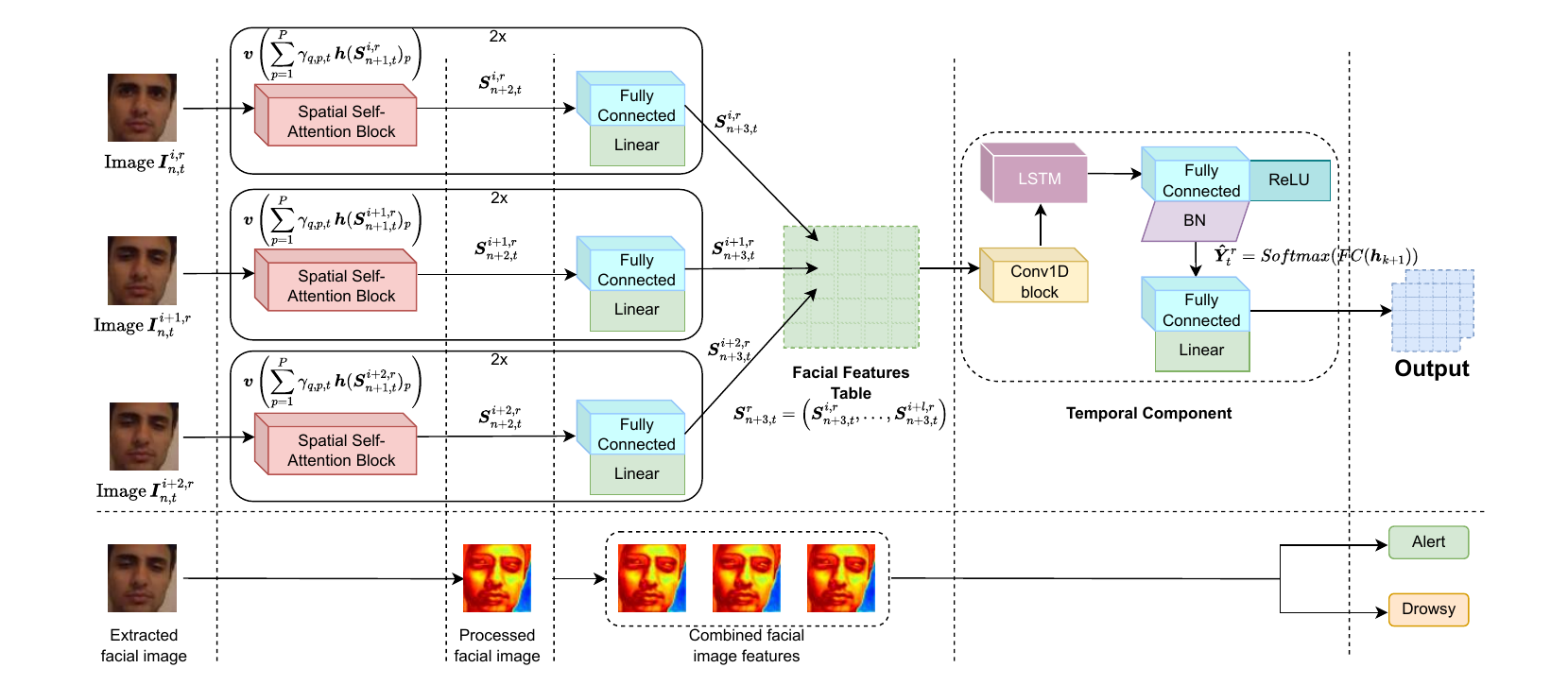}
	\caption{The combination of CNN, SSA, and LSTM to analyze time-series frames.}
	\label{fig:SSTA}
\end{figure*}

The proposed SSTA architecture is described in Fig.~\ref{fig:SSTA}. The images are processed by an \textbf{SSA} block~\cite{kim2020spatially}. They are then converted to vectors by a Fully Connected (FC) Layer with a linear function. After that, continuous images are aggregated into a Facial Features Table. Finally, a Temporal Component with an \textbf{LSTM}~\cite{hochreiter1997long} is used to analyze a series of continuous images to create the output. 
\subsubsection {The SSA Block}
\begin{figure}[t!]
	\includegraphics[width=\linewidth]{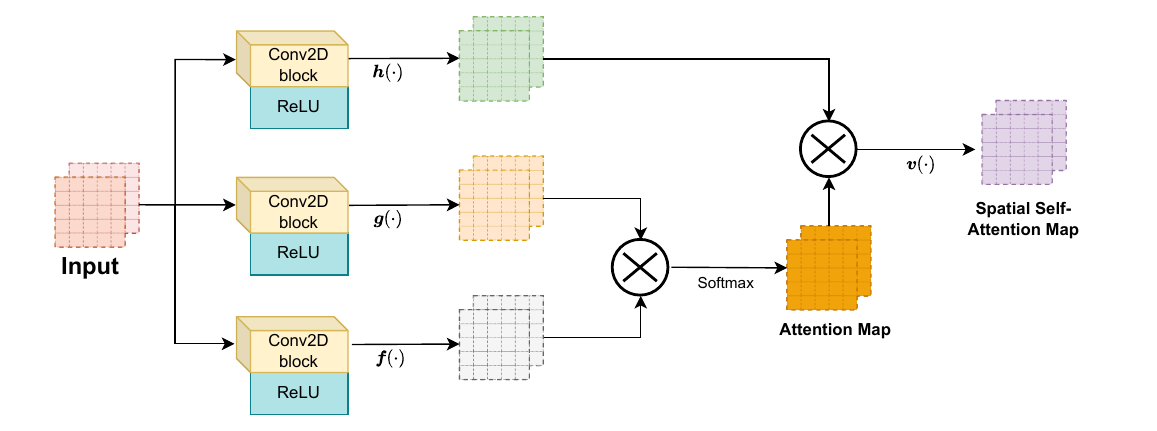}
	\caption{An SSA Block.}
	\label{fig:SSA}
\end{figure}

Fig.~\ref{fig:SSA} describes the SSA block~\cite{kim2020spatially}, which enhances focus on key facial regions in each image, such as the eyes, mouth, and nose. First, a two-dimensional convolutional layer (Conv2D) is applied using a filter bank to extract visual features from the images. Let $\boldsymbol{I}$ denote the set of training images, where $t \in \{1, \dots, T\}$ is the operator index, $i$ is the image index within operator $t$, $r$ is the training iteration, and $n \in \{1, \dots, N\}$ denotes the convolutional layer index. The input image at layer $n$, operator $t$, image $i$, and iteration $r$ is represented as $\boldsymbol{I}_{n,t}^{i,r}$, and the corresponding output feature map is denoted by $\boldsymbol{S}_{n,t}^{i,r}$. The output of convolutional layer $n$ for image $i$ at iteration $r$ is calculated as follows~\cite{saputra2022federated}:
 \begin{equation}
\begin{aligned}
\label{eqn6}
\boldsymbol{S}_{n+1,t}^{i,r} = \theta_{n,t} \Big(\boldsymbol{S}_{n,t}^{i,r} * \boldsymbol{B}_{n,t}\Big),
\end{aligned}
\end{equation}
where $\theta_{n,t}$ is the activation function, $(*)$ is the convolutional operation, and $\boldsymbol{B}_{n,t}$ is the filter bank of layer $n$ in the operator $t$. We then use two functions $\boldsymbol{f}(\cdot)$ and $\boldsymbol{g}(\cdot)$ to transform $\boldsymbol{S}_{n+1,t}^{i,r}$ into two different feature spaces for attention computation with $\boldsymbol{f}(\boldsymbol{S}_{n+1,t}^{i,r})=\boldsymbol{\Omega}_{f}\boldsymbol{S}_{n+1,t}^{i,r}$ and $\boldsymbol{g}(\boldsymbol{S}_{n+1,t}^{i,r})=\boldsymbol{\Omega}_{g}\boldsymbol{S}_{n+1,t}^{i,r}$. Let $P$ denote the total number of spatial positions in the feature map. The attention score between the $q$-th and $p$-th spatial locations is computed by taking the dot product of the corresponding feature vectors from the transformed feature spaces~\cite{zhang2019self}:
\begin{equation}
\begin{aligned}
\label{eqn7_1}
s_{q,p,t} = \boldsymbol{f}(\boldsymbol{S}_{n+1,t}^{i,r})_q^T \, \boldsymbol{g}(\boldsymbol{S}_{n+1,t}^{i,r})_p.
\end{aligned}
\end{equation}

The attention weight $\gamma_{q,p,t}$ is then obtained using the softmax function~\cite{kim2020spatially, zhang2019self}:
\begin{equation}
\begin{aligned}
\label{eqn7}
\gamma_{q,p,t} = \frac{\exp(s_{q,p,t})}{\sum_{p'=1}^{P} \exp(s_{q,p',t})},
\end{aligned}
\end{equation}
where $\gamma_{q,p,t}$ represents how much the model in the operator $t$ attends to the $p$-th location when synthesizing the representation for the $q$-th location. The output of the SSA block with image $i$ of operator $t$ can be calculated as follows~\cite{zhang2019self}: 
\begin{equation}
\begin{aligned}
\label{eqn8}
\boldsymbol{S}_{n+2,t}^{i,r} = \boldsymbol{v} \left( \sum_{p=1}^P \gamma_{q,p,t} \, \boldsymbol{h}(\boldsymbol{S}_{n+1,t}^{i,r})_p \right),
\end{aligned}
\end{equation}
where $\boldsymbol{h}(\cdot)$ and $\boldsymbol{v}(\cdot)$ are learnable transformations defined as $\boldsymbol{h}(\boldsymbol{S}) = \boldsymbol{\Omega}_{h} \boldsymbol{S}$ and $\boldsymbol{v}(\cdot) = \boldsymbol{\Omega}_{v} (\cdot)$.
After this layer, $\boldsymbol{S}_{n+2,t}^{i,r}$ is flattened into a vector by a fully connected layer (FC), denoted as $\boldsymbol{S}_{n+3,t}^{i,r} = \mathrm{FC}(\boldsymbol{S}_{n+2,t}^{i,r})$, where $\mathrm{FC}(\cdot)$ is the function of the fully connected layer. The SSA and FC processes are applied a second time to each image and repeated across $l$ consecutive images.  Afterward, a Facial Features Table is constructed as a sequence of training samples: $\boldsymbol{S}_{n+3,t}^{r} = \left( \boldsymbol{S}_{n+3,t}^{i,r}, \boldsymbol{S}_{n+3,t}^{i+1,r}, \dots, \boldsymbol{S}_{n+3,t}^{i+l,r} \right)$. 

\subsubsection{The Temporal Component with an LSTM}
After the Facial Features Table is generated, a Temporal Component based on LSTM with multiple memory cells is applied to analyze the features across sequential images more effectively~\cite{xia2020lstm}. The LSTM used three weight functions $\boldsymbol{\Phi}_A$, $\boldsymbol{\Phi}_B$, and $\boldsymbol{\Phi}_C$. We denote $\boldsymbol{S}_{n+4,t}^r=\text{Conv1D}(\boldsymbol{S}_{n+3,t}^r)$ as the output of the 1-dimensional convolution layer (Conv1D), $\sigma(\cdot)$ as the sigmoid function, $\phi(\cdot)$ as the tanh function, $\boldsymbol{h}_k$ as the previous output of the LSTM, $\boldsymbol{d}_k$ as the cell at state $k$ of LSTM. The output of LSTM can be calculated as in the following equations~\cite{yu2019review, graves2012long}:
\begin{equation}
\begin{aligned}
\label{eqn9}
&\boldsymbol{a}_k=\sigma(\boldsymbol{\Phi}_{AS} \boldsymbol{S}_{n+4,t}^r + \boldsymbol{\Phi}_{Ah} \boldsymbol{h}_k), \\
&\boldsymbol{b}_k=\sigma(\boldsymbol{\Phi}_{BS} \boldsymbol{S}_{n+4,t}^r + \boldsymbol{\Phi}_{Bh} \boldsymbol{h}_k), \\
&\boldsymbol{c}_k=\phi(\boldsymbol{\Phi}_{CS} \boldsymbol{S}_{n+4,t}^r + \boldsymbol{\Phi}_{Ch} \boldsymbol{h}_k), \\
&\boldsymbol{d}_k=\boldsymbol{d}_{k-1}+\boldsymbol{a}_k \otimes \boldsymbol{c}_k, \\
&\boldsymbol{h}_{k+1}=\boldsymbol{b}_k \otimes \phi(\boldsymbol{d}_k),
\end{aligned}
\end{equation}
where $\otimes$ is the element-wise multiplication of two vectors. At operator $t$, iteration $r$, we denote $\boldsymbol{\hat{Y}}_t^r$ as the final predicted output of the neural network. $\boldsymbol{\hat{Y}}_t^r$ can be calculated as follows:
\begin{equation}
\begin{aligned}
\label{eqn10}
\boldsymbol{\hat{Y}}^r_t=\operatorname{Softmax}(\operatorname{FC}(\boldsymbol{h}_{k+1})).
\end{aligned}
\end{equation}

We denote $\boldsymbol{Y}_t^r$ as the label. We can calculate the loss using the categorical cross-entropy loss function as follows:
\begin{equation}
\begin{aligned}
\label{eqn11}
\mathcal{L}_t^r=-\sum_{p=1}^M \sum_{q=1}^C y_{p,q,t}^r \log(\hat{y}_{p,q,t}^r),
\end{aligned}
\end{equation}
where $M$ is the number of samples, $C$ is the number of classification classes, $y_{p,q,t}^r \in \boldsymbol{Y}_t^r$, and $\hat{y}_{p,q,t}^r \in \boldsymbol{\hat{Y}}_t^r$. Using~\eqref{eqn11}, we can calculate the gradient of the framework of operator $t$ as follows:
\begin{equation}
\begin{aligned}
\label{eqn12}
\nabla \boldsymbol{\beta}_{t}^r = \frac {\partial\mathcal{L}_t^r}{\partial\boldsymbol{w}_t^r} = \frac {\boldsymbol{w}_t^r - \boldsymbol{w}_t^{r-1}}{\mu},
\end{aligned}
\end{equation}
with $\mu$ as the learning rate, $\boldsymbol{w}_t^r$ as the new weight matrix of the neural network of operator $t$ at iteration $r$, $\boldsymbol{w}_t^{r-1}$ as the previous weight matrix of the neural network of operator $t$. Algorithm~\ref{al:SSTA} summarizes this process.
\begin{algorithm}
	\algsetup{linenosize=\tiny}
	\caption{Our Proposed SSTA Framework}
	\label{al:SSTA}
	\begin{algorithmic}[1]
        \FOR{$f \in (i,i+l)$}
            \STATE Operator $t$ uses SSA with functions $\boldsymbol{f}(\cdot), \boldsymbol{g}(\cdot), \boldsymbol{v}(\cdot)$ to calculate $\boldsymbol{S}_{n+2,t}^{f,r}$, 
            \STATE Operator $t$ then uses $\operatorname{FC}(\cdot)$ to calculate $\boldsymbol{S}_{n+3,t}^{f,r}$,      
        \ENDFOR	
        \STATE Operator $t$ uses the results of $l$ image to create $\boldsymbol{S}_{n+3,t}^r$,
        \STATE Operator $t$ then uses LSTM to calculate $\boldsymbol{h}_{k+1}$ and $\boldsymbol{\hat{Y}}_t^r$,
        \STATE Operator $t$ then uses labels and loss function to calculate $\mathcal{L}_t^r$ and $\nabla \boldsymbol{\beta}_{t}^r$.
	\end{algorithmic}
\end{algorithm}


\subsection {Federated Learning Drowsiness Detection}
\begin{figure}[t!]
	\includegraphics[width=\linewidth]{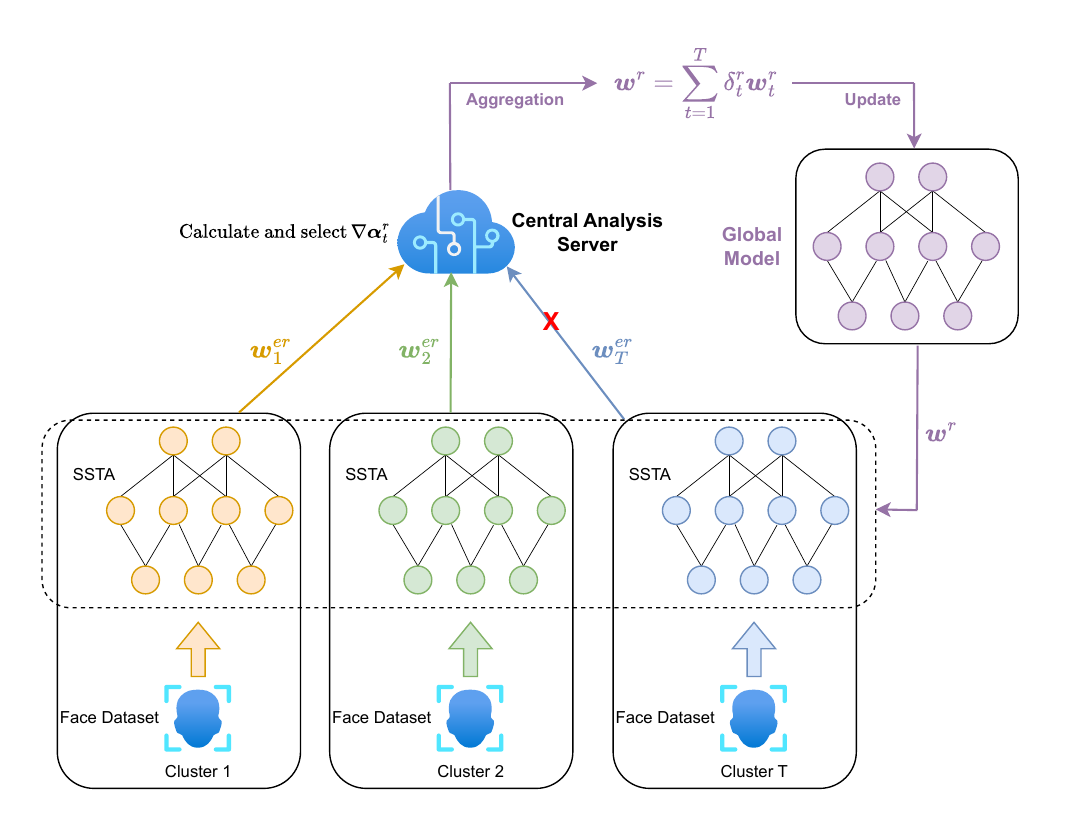}
	\caption{FL with GSC: Each operator sends its trained model to the central analysis server, which performs model selection based on GSC to identify suitable models for aggregation. The selected models are then aggregated to form a new global model.}
	\label{fig:GSFL}
\end{figure}

\subsubsection {Increasing Number of Local Epochs} 
In federated learning, increasing the number of local training epochs in the $FedAvg$ algorithm is essential. Running more local epochs reduces the frequency of communication between operators and the central analysis server, which helps lower the communication overhead~\cite{mcmahan2017communication} and keeps a robust convergence for the global model~\cite{li2020federated}. 
In this paper, we use multiple local epochs within each global round to make our proposed model more robust to data heterogeneity. We denote $e$ as the number of local epochs, the trained model (weight) $\boldsymbol{w}_t^{er}$ is sent from operator $t$ to the central analysis server at global iteration $r$, corresponding to local epochs $er$. In the central analysis server, the gradient $\nabla \boldsymbol{\alpha}_{t}^r$ can be calculated as~\cite{mcmahan2017communication}:
\begin{equation}
\begin{aligned}
\label{eqn13}
\nabla \boldsymbol{\alpha}_{t}^r = \nabla \boldsymbol{\beta}_{t}^{er} = \frac {\boldsymbol{w}_t^{er} - \boldsymbol{w}_t^{er-1}}{\mu}.
\end{aligned}
\end{equation}


\subsubsection {Gradient Similarity Comparison} 
After receiving trained models from operators, the central analysis server calculates a gradient $\nabla \boldsymbol{\alpha}^r$ for each operator as described in Fig.~\ref{fig:GSFL}. It then performs GSC to select suitable gradients to aggregate. To do that, the central analysis server first calculates the pairwise GSC matrix between operators as follows:
\begin{equation}
\begin{aligned}
\label{eqn14}
\boldsymbol{P}_{t,u}^r= \frac{\nabla \boldsymbol{\alpha}_{t}^r \nabla \boldsymbol{\alpha}_{u}^r}{\|\nabla \boldsymbol{\alpha}_{t}^r\| \| \nabla \boldsymbol{\alpha}_{u}^r\|},\quad t,u \in \{1,..., T\}.
\end{aligned}
\end{equation}

It then calculates the average similarity of operator $t$ with other operators as follows:
\begin{equation}
\begin{aligned}
\label{eqn15}
\boldsymbol{\bar{p}}_{t}^r= \frac{1}{T} \sum_{t=1}^T{\boldsymbol{P}_{t,u}^r}.
\end{aligned}
\end{equation}

The central analysis server then creates a threshold to define a group of operators. We denote $\theta$ as the similarity threshold to define a set of valid operators:
\begin{equation}
\begin{aligned}
\label{eqn16}
\Re^r = \{t | \boldsymbol{\bar{p}}_{t}^r \ge \theta\}.
\end{aligned}
\end{equation}

We then can calculate the weights using the softmax function and the temperature parameter $\tau$:
\begin{equation}
\begin{aligned}
\label{eqn17}
\delta_t^r = 
  \begin{cases}
    \frac{\exp(\boldsymbol{\bar{p}}_{t}^r/\tau)}{\sum_{u \in \Re^r} \exp(\boldsymbol{\bar{p}}_{u}^r/\tau)} & \text{if} \quad t \in \Re^r \\
    0 & \text{otherwise}.
  \end{cases}
\end{aligned}
\end{equation}

Fig.~\ref{fig:GSFL} describes this process. In this figure, the central analysis server does not calculate the gradient $\nabla \boldsymbol{\alpha}_{t}$ of operator $t$ when the $\boldsymbol{\bar{p}}_{t}$ of operator $t$ is smaller than $\theta$ ($\boldsymbol{\bar{p}}_{t} <  \theta$). The gradient $\nabla \boldsymbol{\alpha}_{t}$ is ignored when $\boldsymbol{\bar{p}}_t < \theta$ because a low similarity score suggests that the operator’s update is different from the others, possibly due to noise or data drift. This helps ensure that only consistent and reliable gradients are used in the global model aggregation. We then use Equation~\eqref{eqn17} to calculate the final global aggregated weight as follows:
\begin{equation}
\begin{aligned}
\label{eqn18}
\boldsymbol{w}^r = \sum_{t=1}^T \delta_t^r \boldsymbol{w}_t^r,
\end{aligned}
\end{equation}

The central analysis server then sends the final global aggregated weights to all operators to update the operators' neural networks for the next iteration. This process is continuously repeated until reaching a predefined number of iterations. Once this threshold is met, the system produces a final optimized model, which can be deployed across all operators to detect drowsiness in both existing participants and new users who were not part of the initial training. This process is summarized in Algorithm~\ref{al:GSA}.

\begin{algorithm}[t]
	\algsetup{linenosize=\tiny}
	\caption{Our Proposed Drowsiness Detection Framework}
	\label{al:GSA}
	\begin{algorithmic}[1]
	    \WHILE{$r$ $\leq$ maximum number of iterations}
	        \FOR{$\forall t \in T$}
        		\STATE Operator $t$ uses SSA and LSTM calculate $\boldsymbol{\hat{Y}}_t^r$ as in Equation~\eqref{eqn10}, 
        		\STATE Operator $t$ uses $\boldsymbol{\hat{Y}}_t^r$, its labels $\boldsymbol{Y}_t^r$, and categorical cross-entropy loss function to update $\boldsymbol{w}_t^{er}$,
        		\STATE Operator $t$ sends $\boldsymbol{w}_t^{er}$ to the central analysis server.		        
            \ENDFOR	
            \STATE The central analysis server uses $\boldsymbol{w}_t^{er}$ from operators to calculate $\nabla \boldsymbol{\alpha}^r$ as in Equation~\eqref{eqn13} and $\boldsymbol{\bar{p}}^r$ as in Equation~\eqref{eqn15} for each operator,
            \STATE The central analysis server calculates $\Re^r$ as in Equation~\eqref{eqn16} and $\delta_t^r$ as in Equation~\eqref{eqn17}.
            \STATE The central analysis server calculates the final aggregated weight $\boldsymbol{w}^r$ as in Equation~\eqref{eqn18}.
            \STATE The central analysis server sends $\boldsymbol{w}^r$ to all operators to update their neural network.
            \STATE {$r=r+1$.}
		\ENDWHILE
		\STATE Operators use the optimized model to detect drowsiness.
	\end{algorithmic}
\end{algorithm}


\section{Experiment setup}\label{sec:exp}

\subsection{Dataset}
In this paper, we use the University of Texas at Arlington Real-Life Drowsiness Dataset (UTA-RLDD)~\cite{ghoddoosian2019realistic, UTA_dataset} to evaluate the performance of our proposed framework in comparison with other state-of-the-art models. This dataset was developed by the University of Texas for multi-stage drowsiness detection. The UTA-RDD dataset contains approximately 30 hours of RGB video recordings, totaling 180 videos from 48 healthy participants. Each participant contributed three videos, one for each of the three classes: alertness, low vigilance, and drowsiness. The videos were recorded from various angles in diverse real-world settings and backgrounds. All videos were recorded at an angle that ensured both eyes were visible, with the camera positioned within arm’s length of the participant. These instructions were designed to make the recordings resemble videos captured in a car, where a phone is placed in a dashboard-mounted holder while driving. Each participant self-recorded their videos using either a smartphone or a webcam. The frame rate remained below 30 fps, aligning with the typical frame rates of consumer-grade cameras.

\subsection{Evaluation Method}
The confusion matrix~\cite{confusion_matrix1, confusion_matrix2} is widely used to evaluate machine learning models and is particularly effective in assessing the performance of drowsiness detection~\cite{fu2024survey}. In this context, TP, TN, FP, and FN represent True Positive, True Negative, False Positive, and False Negative, respectively. In this paper, we evaluate model performance using key metrics derived from the confusion matrix, namely accuracy, precision, and recall. The accuracy of a model is calculated as follows:
\begin{equation}
\begin{aligned}
\label{eqn19}
	\mbox{Accuracy} = \frac{\mbox{TP}+\mbox{TN}}{\mbox{TP}+\mbox{TN}+\mbox{FP}+\mbox{FN}}.
\end{aligned}
\end{equation}

In addition, we use precision and recall to evaluate the performance of the models. Given $B$ as the number of classification groups (i.e., alert and drowsy), the precision is calculated as follows:
\begin{equation}
\begin{aligned}
\label{eqn20}
	\mbox{Precision} = \sum_{b=1}^B\frac{\mbox{TP}_b}{\mbox{TP}_b+\mbox{FP}_b}.
\end{aligned}
\end{equation}

Similarly, the recall of the system can be calculated as follows:
\begin{equation}
\begin{aligned}
\label{eqn21}
	\mbox{Recall} = \sum_{b=1}^B\frac{\mbox{TP}_b}{\mbox{TP}_b+\mbox{FN}_b}.
\end{aligned}
\end{equation}

\subsection{Simulation Setup}
As described above, the UTA-RLDD dataset includes 48 participants, each participant is categorized into 3 classes corresponding to drowsiness levels: alertness, low vigilance, and drowsiness. To demonstrate that our proposed model can efficiently work with different individuals without any prior training, we divide the participants into training and testing datasets. The training dataset includes 42 participants. 
The testing dataset consists of 12 participants, including 6 from the training dataset and 6 who are not included in the training data. These 6 unseen participants are used to evaluate the model’s performance on individuals it has not encountered before.
The training dataset is split with 80\% of the data for training and 20\% for validation. The validation data is used to evaluate the accuracy and the convergence of the training process. In our experiment, our preprocessing tool first extracts frames from the training videos. After that, it performs face detection and extraction, generating a dataset containing 292,220 frames. Due to the high computational cost and time required to process the entire dataset, we randomly use 50\% of the generated frames to perform experiments.

We consider two different scenarios, including centralized and federated learning. The centralized learning scenario is considered as the benmarks for our proposed FL model. In this scenario, all participants are used to train the machine learning models to evaluate performance. In contrast, in federated learning scenarios, the participants are grouped into different operators. Each operator includes a distinct set of participants, i.e., in the case of five operators, each operator consists of eight participants, whereas in the case of 42 operators, each operator contains a single participant. We perform experiments using five workstations running the Ubuntu operating system, each equipped with a GPU. The setup includes two NVIDIA GeForce RTX 3090, two NVIDIA RTX 6000 Ada Generation, three NVIDIA A100, and one NVIDIA GeForce RTX 4090 graphics cards, using the PyTorch framework for computational tasks.


\section{Performance Evaluation}
We compare our proposed framework with other state-of-the-art models to demonstrate the outperformance of our proposed framework. We consider two different scenarios, including the centralized and federated learning models. 

\begin{figure}
    \centering
    \includegraphics[width=\linewidth]{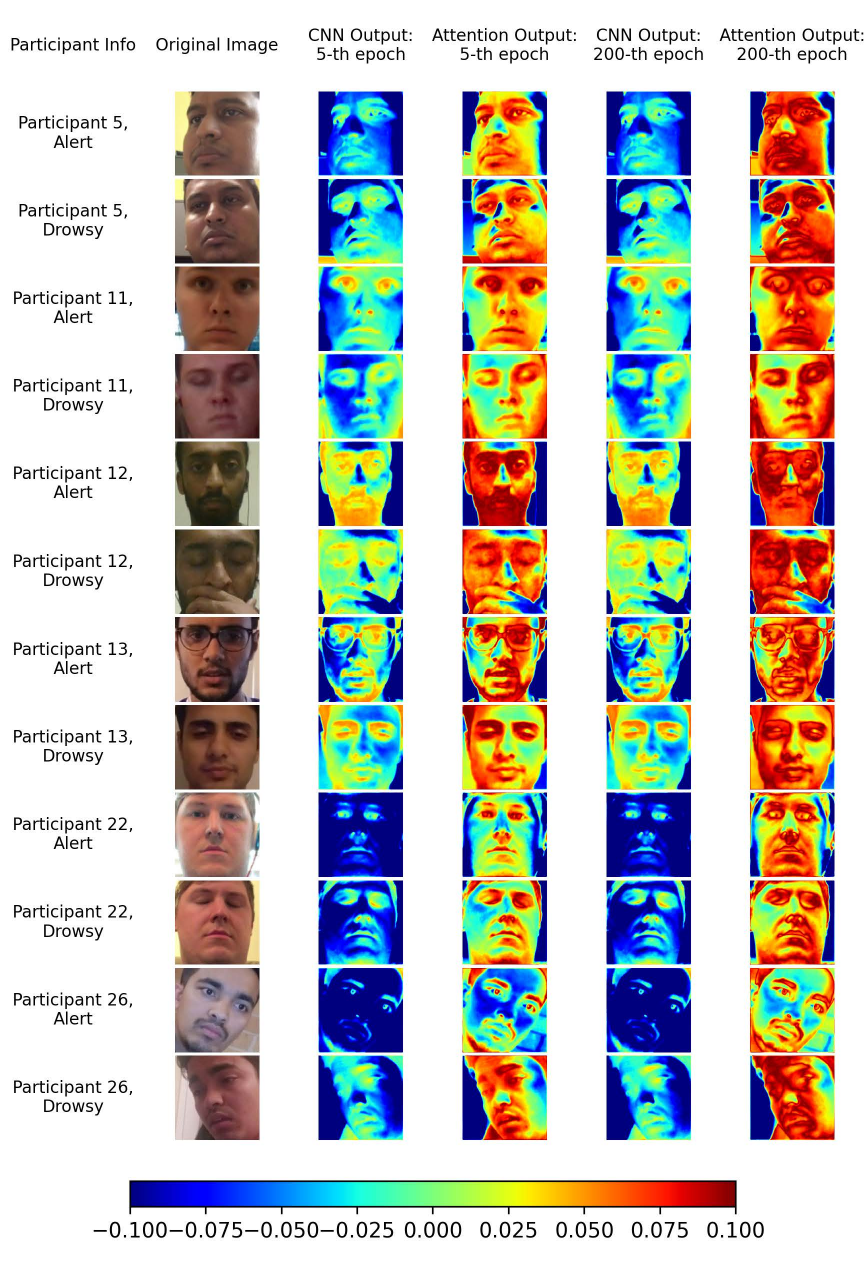}
    \caption{Visualization of the SSA mechanism's output across selected participants in the testing dataset.}
    \label{fig:visualization}
\end{figure}

\subsection{Centralized Learning Evaluation}

In the centralized learning scenario, we consider a scenario in which a centralized server can be used to collect all training datasets from the operator to perform a centralized training process. This serves as a baseline to compare our proposed approach with other state-of-the-art machine learning techniques. In particular, we
 evaluate the performance of our proposed SSTA model in comparison with other state-of-the-art framework such as Vision Transformer (ViT)~\cite{Bin2024}, LSTM~\cite{liu2021effects}, Convolutional Neural Network (CNN)~\cite{Tamanani2021}, Multilayer Perceptron (MLP)~\cite{Mohammadi2023}, Decision Tree (DT)~\cite{mittal2021driver}, K-Nearest Neighbor  (KNN)~\cite{mittal2021driver}, Linear Regression (LR)~\cite{mittal2021driver}, Random Forest (RF)~\cite{mittal2021driver}, Support Vector Machine (SVM)~\cite{mittal2021driver}, and Extreme Gradient Boosting (XGBoost)~\cite{mittal2021driver}. 

\subsubsection{Visualization of SSA Patterns}
Fig.~\ref{fig:visualization} shows the output of the CNN baseline and our proposed model after the SSA blocks at epoch 5 (early stage of training) and epoch 200 (late stage of training), across 12 participants from the testing dataset. To ensure diverse testing conditions, we select participants with varying characteristics, such as participant 13 wears glasses, participant 25 whose videos are captured at a large rotation angle, participant 12 covers his mouth when feeling drowsy, participants 5 and 11 are in low-light environments while in a drowsy state, and participants 12 and 22 use headphones.  
Compared to the CNN, our model produces stronger activations, represented by orange to red colors, around key facial areas such as the eyes and mouth at both training stages. These visualizations indicate that our model can be more effective in focusing on important facial regions (eyes, nose, and mouth), enhancing its ability to detect signs of drowsiness throughout the training process.

\subsubsection{Model Accuracy and Convergence Comparison}
\begin{figure}
    \centering
    \includegraphics[width=.48\textwidth]{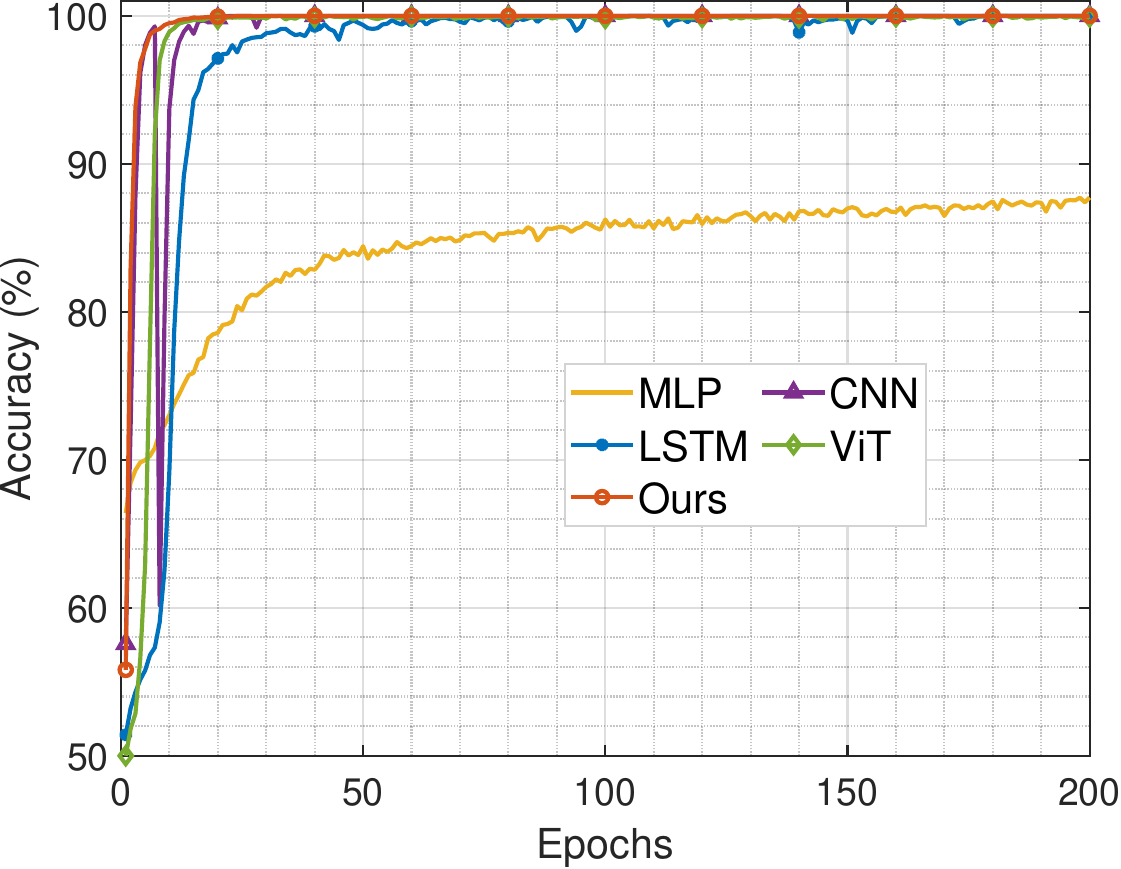}
    \caption{Training accuracies of different centralized learning approaches evaluated on participants from the training dataset.}
    \label{fig:cl}
\end{figure}

Fig.~\ref{fig:cl} describes the convergence and accuracy of the training process for our proposed SSTA model compared to other DL models. As observed, both our proposed model and the ViT achieve convergence within the first 20 epochs. Notably, during the training process, our proposed SSTA model converges more rapidly and attains the highest accuracy (100\%) compared to the others. Although both CNN and ViT models achieve near-perfect accuracy (100\%) after 20 epochs, the CNN experiences a significant drop in accuracy around epoch 10, while the MLP stabilizes at approximately 80\% accuracy after 100 epochs of the training process.
Table~\ref{tab:perf_cl} shows the performance results of our proposed SSTA model in terms of accuracy, precision, and recall, compared to other models on the testing dataset, which includes 6 previously untrained and 6 other trained participants. As shown in the table, traditional machine learning models, i.e., DT, KNN, LR, RF, SVM, XGBoost, exhibit relatively poor performance, achieving accuracies ranging from approximately 60\% to 77\% on the testing dataset. MLP also demonstrates suboptimal performance, with an accuracy of 71.3\%. In contrast, deep learning models such as CNN, LSTM, and ViT perform significantly better, achieving accuracies of 83.92\%, 90.38\%, and 89.32\%, respectively. Notably, with the testing dataset, our proposed SSTA model (Ours) outperforms all other models by getting the highest accuracy, precision, and recall of 91.83\%, 91.96\%, and 91.83\%, respectively.


\begin{table*}
\centering
\footnotesize
\caption{Centralized mode: Performance comparison on testing dataset, including 6 trained and 6 untrained participants.}
\label{tab:perf_cl}
\begin{tblr}{
  width = \linewidth,
  colspec = {Q[80]Q[70]Q[70]Q[70]Q[70]Q[70]Q[110]Q[70]Q[70]Q[85]Q[70]Q[70]},
  column{even} = {c},
  column{3} = {c},
  column{5} = {c},
  column{7} = {c},
  column{9} = {c},
  column{11} = {c},
  cell{1}{2-12} = {c},
  hlines,
  vlines,
}
~ & \textbf{DT}~\cite{mittal2021driver} & \textbf{KNN}~\cite{mittal2021driver} & \textbf{LR}~\cite{mittal2021driver} & \textbf{RF}~\cite{mittal2021driver} & \textbf{SVM}~\cite{mittal2021driver} & \textbf{XGBoost}~\cite{mittal2021driver} & \textbf{MLP}~\cite{Mohammadi2023} & \textbf{CNN}~\cite{Tamanani2021} & \textbf{LSTM}~\cite{liu2021effects} & \textbf{ViT}~\cite{Bin2024} & \textbf{Ours}\\
\textbf{Accuracy} & 71.1254 & 77.8734 & 61.2936 & 77.9235 & 64.5475 & 72.6672 & 71.3056 & 83.9277 & 90.3805 & 89.3219 & \textbf{91.8341}\\
\textbf{Precision} & 71.3131 & 77.8845 & 61.4571 & 77.9756 & 64.9342 & 72.6910 & 71.3168 & 83.9819 & 90.3826 & 89.3913 & \textbf{91.9607}\\
\textbf{Recall} & 71.1254 & 77.8734 & 61.2936 & 77.9235 & 64.5475 & 72.6672 & 71.3056 & 83.9277 & 90.3805 & 89.3219 & \textbf{91.8341}
\end{tblr}
\end{table*}

\subsection{Federated Learning Evaluation}
\begin{table*}
\centering
\footnotesize
\caption{FL mode with 5 operators: Performance comparison on testing dataset, including 6 trained and 6 untrained participants.}
\label{tab:perf_fl}
\begin{tblr}{
  width = .8\linewidth,
  colspec = {Q[102]Q[150]Q[150]Q[150]Q[150]Q[200]Q[100]},
  column{even} = {c},
  column{3} = {c},
  column{5} = {c},
  column{7} = {c},
  cell{1}{2-7} = {c},
  hlines,
  vlines,
}
 & \textbf{CNN\_FedAvg} & \textbf{LSTM\_FedAvg} & \textbf{ViT\_FedAvg} & \textbf{SSTA\_FedAvg} & \textbf{SSTA\_FedProx}~\cite{li2020federated} & \textbf{Ours}\\
\textbf{Accuracy} & 76.5521 & 80.8601 & 81.2739 & 74.9849 & 85.1201 & \textbf{89.9253}\\
\textbf{Precision} & 77.5237 & 82.5897 & 81.2739 & 76.4987 & 85.1376 & \textbf{90.6377}\\
\textbf{Recall} & 76.5521 & 80.8601 & 81.2739 & 74.9849 & 85.1201 & \textbf{89.9253}
\end{tblr}
\end{table*}

In this section, we consider a decentralized federated learning setting in which participants are organized into operators for the training process. Each operator may contain one or more participants. To evaluate the impact of clustering, we perform experiments with 5, 10, and 42 operators. A total of 42 participants in the training dataset are randomly and equally assigned to the operators. This results in 8 participants per operator for 5 operators, 4 participants per operator for 10 operators, and 1 participant per operator when 42 operators are used.

\subsubsection{Comparative Analysis of Federated Strategies}
\begin{figure}
    \centering
    \includegraphics[width=.48\textwidth]{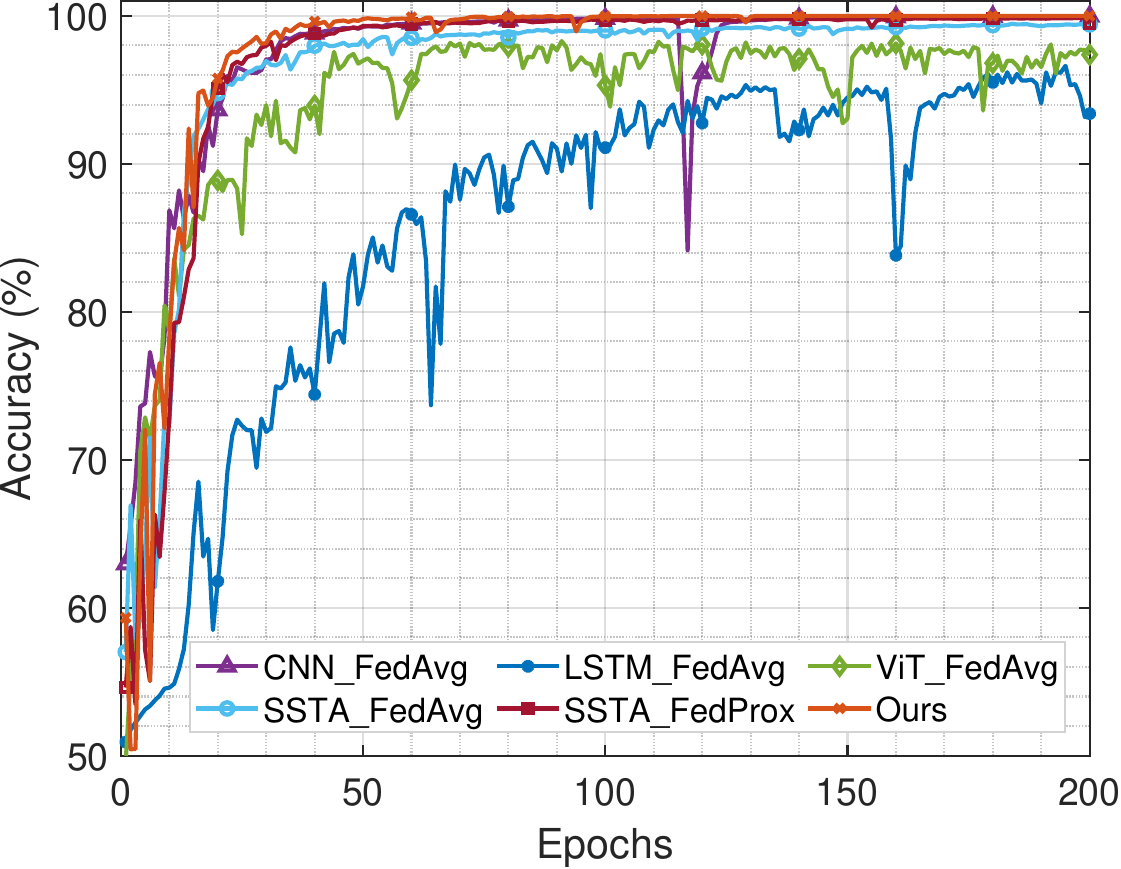}
    \caption{The accuracies of different FL models in the case of 5 operators during the training process.}
    \label{fig:fl_5}
\end{figure}

Fig.~\ref{fig:fl_5} shows the accuracy curves during the training process of different combinations of deep learning and federated learning approaches with 5 operators. The approaches include CNN-FedAvg, LSTM-FedAvg, ViT-FedAvg, as well as SSTA-FedAvg, SSTA-FedProx, and our proposed approach: the complete SSTA with GSC (SSTA-GS). All approaches eventually reach convergence; however, LSTM-FedAvg and ViT-FedAvg show noticeable fluctuations, indicating their instability during the training process. Although CNN-FedAvg appears to converge, it experiences a significant drop in accuracy around epoch 118, raising concerns about its reliability. In contrast, our proposed SSTA-based approaches, when combined with FedAvg, FedProx, or GSC, achieve more stable and consistent convergence, suggesting improved training performance in the 5 operators setting.

\begin{figure*}[!t]
    \centering
    \begin{subfigure}[b]{\textwidth}
        \centering
        \includegraphics[width=\textwidth]{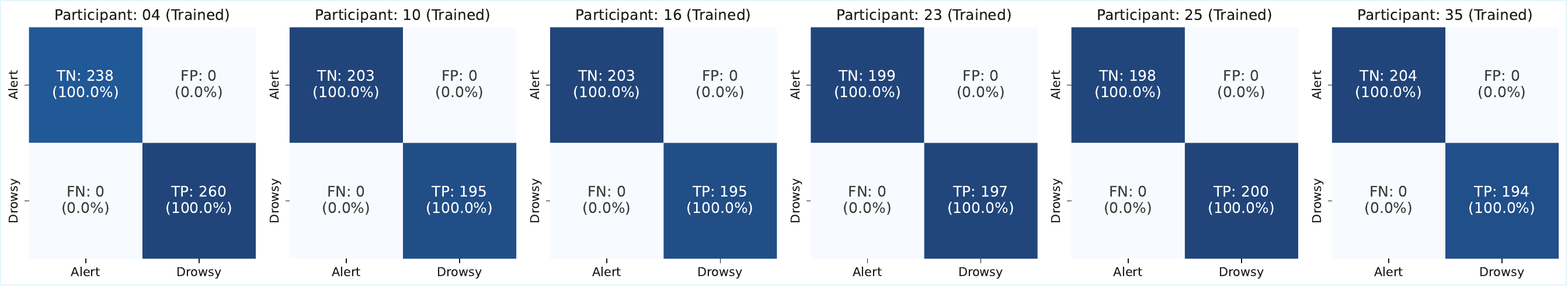}
        \caption{Trained participants}
    \end{subfigure}
    \hfill
    \begin{subfigure}[b]{\textwidth}
        \centering
        \includegraphics[width=\textwidth]{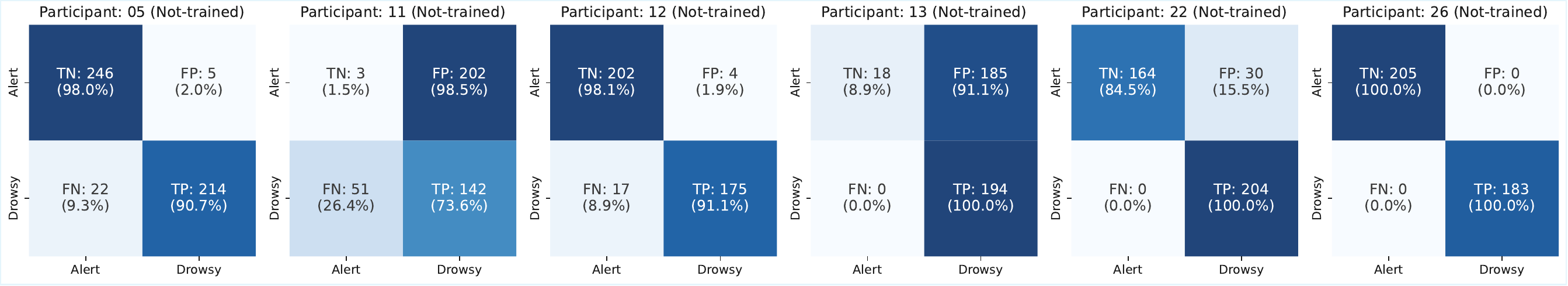}
        \caption{Not-trained participants}
    \end{subfigure}
    \caption{Classification report of FL with 5 operators on testing dataset (6 trained and 6 untrained participants).}
    \label{fig:cfm}
\end{figure*}

Table~\ref{tab:perf_fl} presents the performance of the different approaches on the testing dataset. Our proposed approach achieves the highest results, with an accuracy of 89.92\%, precision of 90.63\%, and recall of 89.92\%. These results are approximately 4-5\% higher than those of SSTA\_FedProx across all three metrics and outperform the remaining approaches. These results show that our model is more effective at learning relevant patterns from the heterogeneous data, suggesting its strong potential for real-world applications in federated learning settings. Fig.~\ref{fig:cfm} shows the classification results of our proposed model using federated learning with 5 operators on the testing dataset. 
Our model is 100\% accurate in detecting drowsiness among participants who are part of the training dataset. The model also performs well in classifying previously unseen participants, detecting drowsiness with 100\% accuracy in three participants and more than 90\% accuracy in two participants. The model, however, achieves a lower accuracy of 73\% in detecting drowsiness in the remaining participant, which may be due to differences in data patterns that are not well represented during training. 
This highlights the need for more diverse training data or personalized adjustments to improve performance for all users.

\begin{figure}
    \centering
    \includegraphics[width=.48\textwidth]{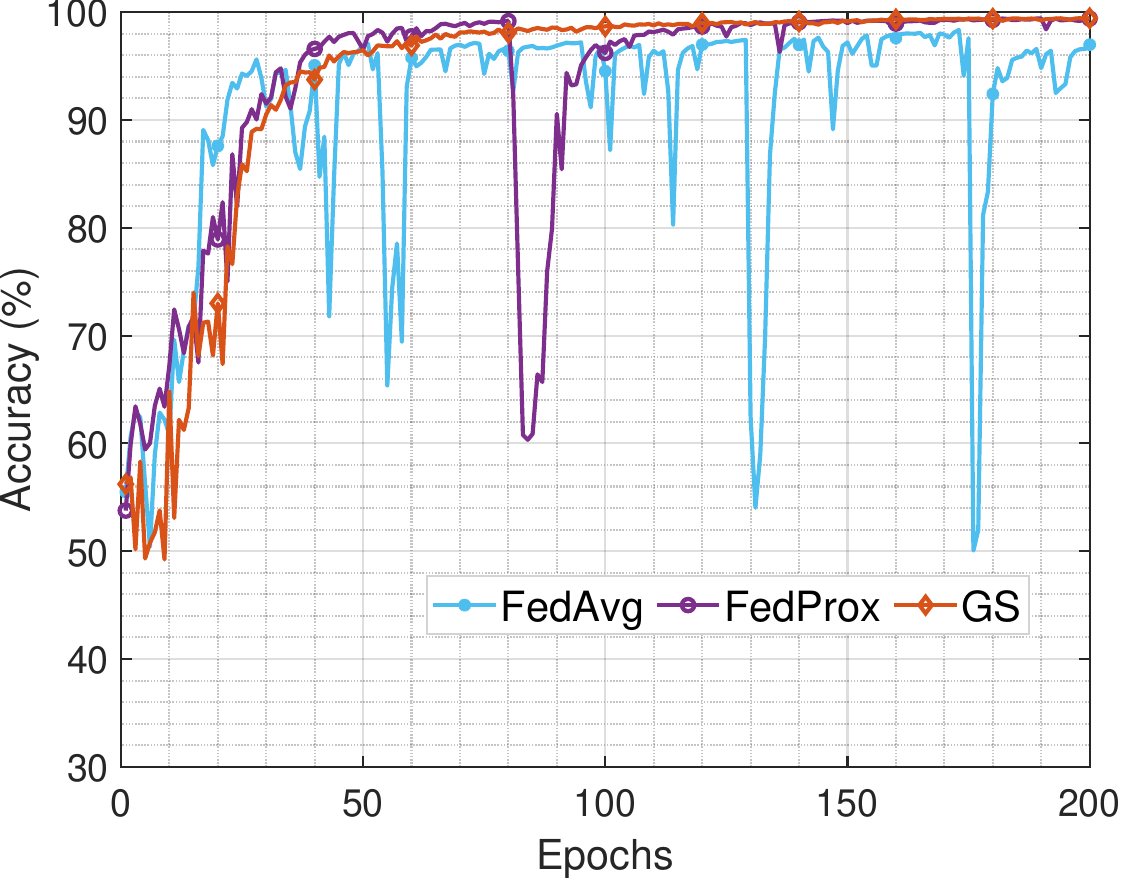}
    \caption{The stability during the training process of different aggregation schemes on our SSTA network with 10 operators.}
    \label{fig:aggregations}
\end{figure}

Fig.~\ref{fig:aggregations} shows the accuracy of different federated learning algorithms using the SSTA model during the training process. We compare three approaches: FedAvg, FedProx, and our proposed approach under a scenario with 10 operators. When the number of operators increases from 5 to 10, data heterogeneity also increases, posing significant challenges for federated learning. This increased heterogeneity directly affects the stability and convergence of the training process, as reflected in the accuracy trends over time. In particular, the FedAvg method shows considerable fluctuations in accuracy throughout training, indicating its instability when handling non-IID data across multiple operators. Besides, FedProx, designed to address some of these issues, achieves a more stable performance than FedAvg but still experiences a sharp drop in accuracy around epoch 85, highlighting its limitations under high heterogeneity. In contrast, our proposed approach consistently outperforms both baselines in terms of accuracy and stability. The training curve of our proposed model is smooth and stable, demonstrating strong convergence behavior even under challenging conditions. This result highlights the robustness of our method in handling heterogeneous data distributions and maintaining reliable performance across a larger number of operators. These findings confirm that our approach offers a more effective and resilient solution for federated learning in realistic, non-IID environments.

\subsubsection{Robustness to Varying Data Heterogeneity}
\begin{figure}
    \centering
    \includegraphics[width=.48\textwidth]{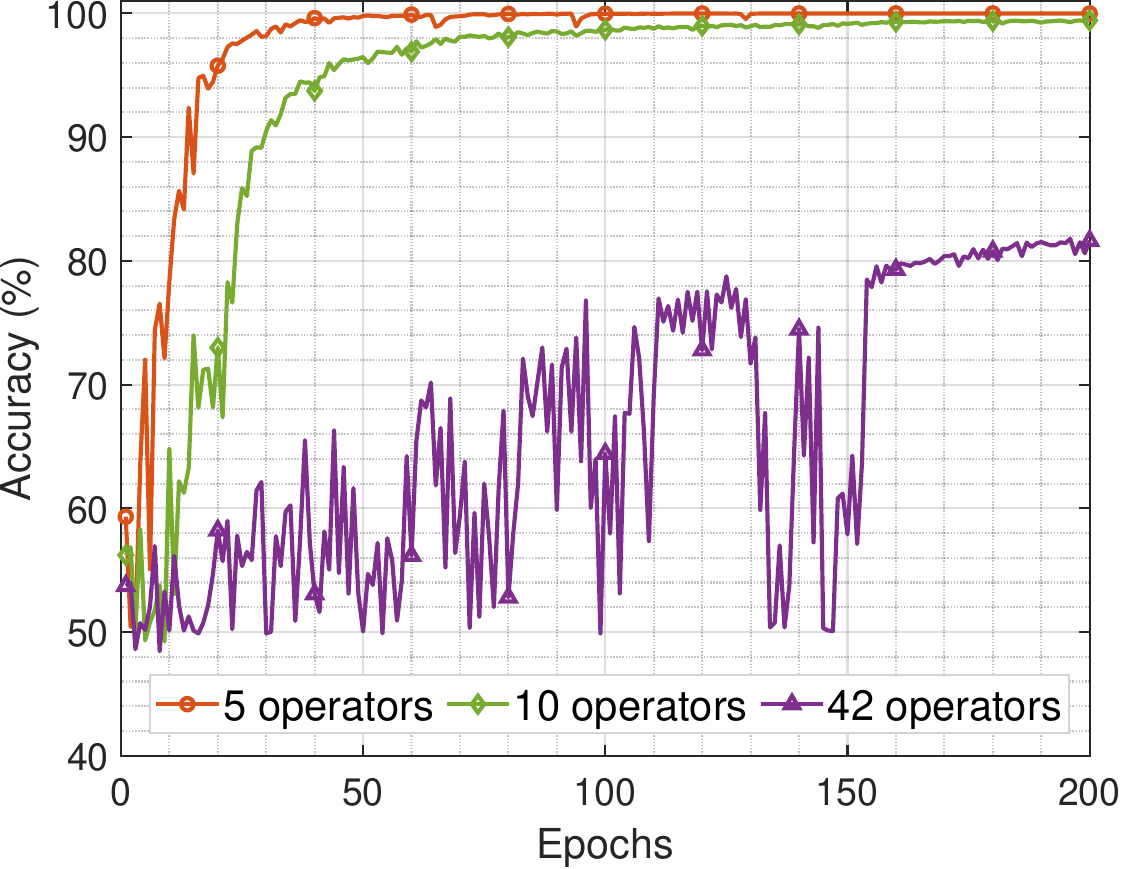}
    \caption{The accuracies of our proposed model in the training process when the number of operators varies.}
    \label{fig:clusters}
\end{figure}

As mentioned in the previous section, increasing the number of operators also increases data heterogeneity. Fig.~\ref{fig:clusters} illustrates the accuracy curves when using different numbers of operators with our proposed model during the training process. We observe that FL with 5 operators converges the fastest, followed by 10 operators. FL with 5 operators reaches 100\% accuracy after approximately 50 epochs. In comparison, FL with 10 operators takes around 100 epochs to converge, while FL with 42 operators (i.e., one participant per operator) requires about 160 epochs. This demonstrates the increased complexity of training under highly heterogeneous data conditions. While increasing the number of operators leads to greater data heterogeneity, our model remains effective in handling these variations, achieving over 80\% training accuracy after 160 epochs, despite a slower convergence rate in comparison with other cases.

Table~\ref{tab:clusters} presents the performance in terms of accuracy, precision, and recall for different operator settings with our proposed model on the testing dataset. FL with 5 operators achieves the highest performance, with an accuracy of 89.92\%, precision of 90.63\%, and recall of 89.92\%. Even in the most heterogeneous case with 42 operators, the model still performs well, achieving 83\% accuracy, 84.47\% precision, and 83.02\% recall. These results highlight the robustness of our proposed model across varying levels of data heterogeneity. The model’s strong and consistent performance demonstrates its suitability for real-world federated learning scenarios. Moreover, the relatively small performance drop in testing accuracy in highly heterogeneous conditions suggests that the model effectively mitigates the adverse effects of non-IID data distributions, which is a common challenge in practical FL applications.
\begin{table}
\small
\centering
\caption{The stabilities of accuracies in the testing dataset when the number of operators increases.}
\label{tab:clusters}
\begin{tblr}{
  width = \linewidth,
  colspec = {Q[171]Q[200]Q[200]Q[200]},
  cell{1-4}{2-4} = {c},
  hlines,
  vlines,
}
~ & \textbf{5 operators} & \textbf{10 operators} & \textbf{42 operators}\\
\textbf{Accuracy} & \textbf{89.9253} & 88.6332 & 83.0204\\
\textbf{Precision} & \textbf{90.6377} & 88.6692 & 84.4740\\
\textbf{Recall} & \textbf{89.9253} & 88.6332 & 83.0204
\end{tblr}
\end{table}


\subsubsection{Impact of Local Epochs on FL Performance}
\begin{figure}
    \centering
    \includegraphics[width=.48\textwidth]{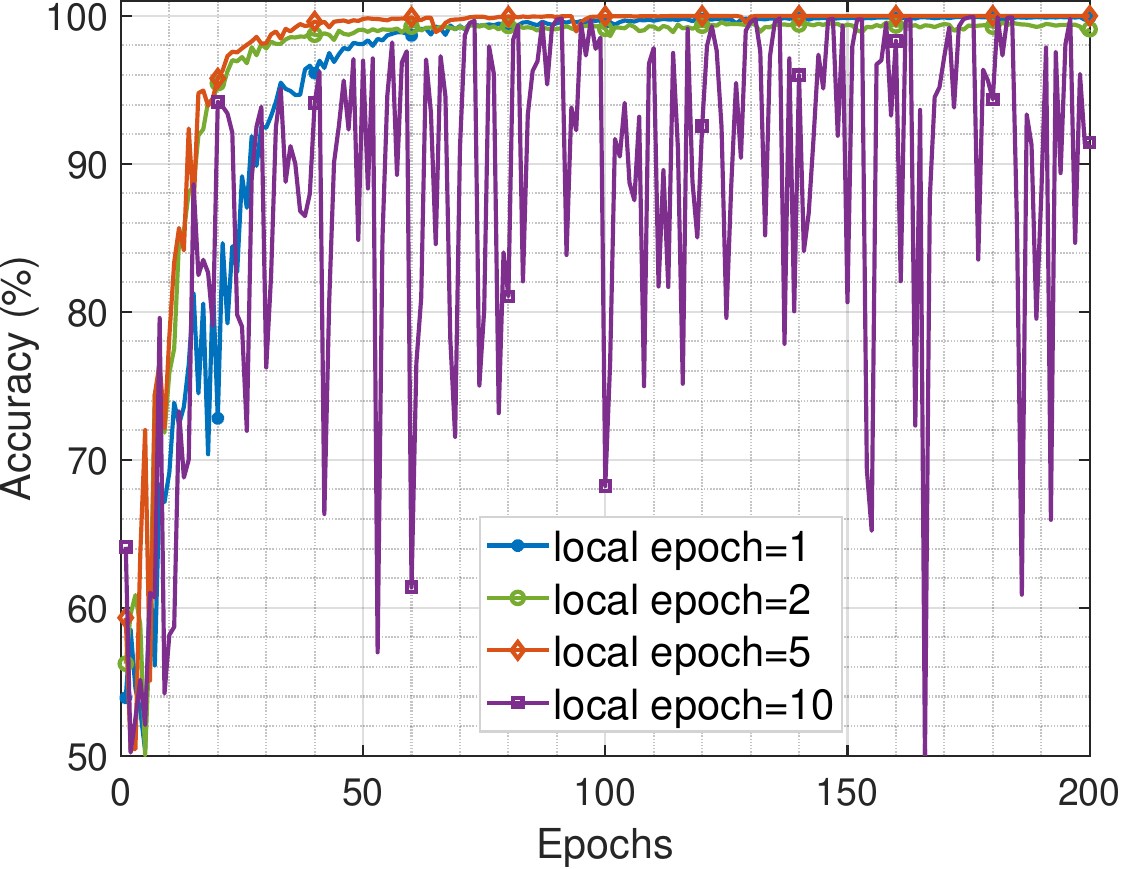}
    \caption{The accuracies of the proposed SSTA model in the training process while varying the number of local epochs.}
    \label{fig:localep}
\end{figure}

The number of local epochs plays a critical role in determining the accuracy and convergence behavior of federated learning (FL), particularly in decentralized environments with heterogeneous data distributions~\cite{li2020federated}. Increasing the number of local epochs allows operators to perform more local updates before communicating with the central analysis server, which can be beneficial in reducing communication overhead and improving model personalization~\cite{mcmahan2017communication}. However, when local datasets are highly heterogeneous, excessively increasing the number of local epochs can lead to overfitting on local data and cause operator models to converge to local optima. This divergence in local updates may negatively impact the global model's ability to converge effectively. Nevertheless, prior studies have shown that selecting an appropriate number of local epochs can improve overall model performance while maintaining stable convergence~\cite{li2020federated}.

In this section, we evaluate the impact of varying the number of local epochs on the stability of the training process. Fig.~\ref{fig:localep} illustrates the effects of different local epoch settings (1, 2, 5, and 10) on training performance in a federated learning scenario. The curve corresponding to 1 local epoch demonstrates slower convergence, likely due to insufficient local training that leads to high communication frequency with limited improvement per round. In contrast, the model with 5 local epochs converges significantly faster and achieves superior performance, indicating an effective balance between local update depth and global model alignment. On the other hand, using 10 local epochs introduces instability and noticeable fluctuations in the learning curve, suggesting overfitting to local data and difficulty in achieving coherent global aggregation.

Based on these empirical results, we adopt 5 local epochs as the optimal setting for all experiments in this study. This choice offers a practical trade-off between communication efficiency, model convergence speed, and global accuracy in the presence of non-IID data.


\section{Conclusion}\label{sec:conclusion}

In this paper, we developed a novel framework for driver drowsiness detection in decentralized environments with heterogeneous facial data. To improve detection accuracy, we combined an SSA mechanism with an LSTM network, helping the model focus on important facial features across different individuals. Moreover, we integrated the GSC into our model to support federated learning, allowing the system to select and combine models from similar client clusters. This improves both the accuracy and robustness of the final global model. Additionally, we built a preprocessing tool that can perform frame extraction from videos, face detection and extraction, and frame augmentation to enhance the data quality of the dataset. Extensive simulations demonstrate that our approach outperforms existing methods in both accuracy and computational efficiency. Furthermore, by enabling decentralized learning without compromising performance, our framework enhances data privacy, making it well-suited for real-world applications in intelligent transportation systems.


\bibliographystyle{IEEEtran}
\bibliography{references}

\end{document}